\documentclass[letterpaper]{article} 
\usepackage{aaai23}  
\usepackage{times}  
\usepackage{helvet}  
\usepackage{courier}  
\usepackage[hyphens]{url}  
\usepackage{graphicx} 
\usepackage{natbib}  
\usepackage{caption} 
\usepackage{algorithm}
\usepackage{algorithmic}
\usepackage{subfigure}
\usepackage{multirow}
\usepackage{amsmath}
\usepackage{graphicx}
\usepackage{stfloats}
\usepackage{multirow}
\usepackage{listings}
\usepackage{xcolor}
\usepackage{amsmath}
\usepackage{subfigure}

\definecolor{codegreen}{rgb}{0,0.6,0}
\definecolor{codegray}{rgb}{0.5,0.5,0.5}
\definecolor{codepurple}{rgb}{0.58,0,0.82}
\definecolor{backcolour}{rgb}{0.95,0.95,0.92}
\usepackage{newfloat}
\usepackage{listings}
\DeclareCaptionStyle{ruled}{labelfont=normalfont,labelsep=colon,strut=off} 
\floatstyle{ruled}
\newfloat{listing}{tb}{lst}{}
\floatname{listing}{Listing}
\newtheorem{myexample}{Example}

\title{AAAI Press Anonymous Submission\\Instructions for Authors Using \LaTeX{}}

\title{Neural Feature-Adaptation for Symbolic Predictions Using Pre-Training and Semantic Loss}
\author {
    Vedant Shah\textsuperscript{\rm *,1,2},
    Aditya Agrawal \textsuperscript{\rm *,1},
    Lovekesh Vig \textsuperscript{\rm 3},
    Ashwin Srinivasan\textsuperscript{\rm 1},
    Gautam Shroff\textsuperscript{\rm 3},
    Tanmay Verlekar\textsuperscript{\rm 1}
}
\affiliations {
    \textsuperscript{\rm *} Equal Contribution\\
    \textsuperscript{\rm 1} APPCAIR, Birla Institute of Technology and Science, Pilani\\
    \textsuperscript{\rm 2} Mila, University of Montreal\\
    \textsuperscript{\rm 3} Tata Consultancy Services\\
    \texttt{vedantshah2012@gmail.com, f20201381@goa.bits-pilani.ac.in}
}

\usepackage{bibentry}

\begin{document}

\maketitle

\begin{abstract}
We are interested in neuro-symbolic systems consisting
of a high-level symbolic layer for explainable prediction in terms
of human-intelligible concepts; and a
low-level neural layer for extracting symbols required
to generate the symbolic explanation. Unfortunately real data is often imperfect. This means that even if the symbolic theory
remains unchanged, we may still need to address the problem of mapping raw data to
high-level symbols, each time there is a
change in the data acquisition environment or equipment
(for example, the clinical explanation of a heart arrhythmia
could be unchanged, but the raw data could vary from
one hospital to another). Manual (re-)annotation
of the raw data each time this happens is laborious and expensive; and automated labelling methods are often imperfect, especially for complex problems.
Recently, the NEUROLOG system proposed the use of
a semantic loss function
 that allows an existing feature-based symbolic model 
to guide the extraction of feature-values from raw data, using a mechanism
called `abduction'. However, the experiments demonstrating the
use of semantic loss through abduction appear to rely heavily
on a domain-specific pre-processing step that enables a prior delineation of
feature locations in the raw data. In this paper, we examine the use of
semantic loss in domains where such pre-processing is not possible, or is not obvious. Using controlled
 experiments on two simulated datasets, we show that without any prior information about
the features, the NEUROLOG approach can continue to predict accurately even with
substantially incorrect
feature predictions (that is, predictions are correct, but explanations
are wrong). We show also that prior
information about the features in the form of
(even imperfect) pre-training can help correct
this situation. These findings are replicated on the original
problem considered by NEUROLOG, without the use of
feature-delineation. This suggests that symbolic explanations
constructed for data in a domain could be re-used in a related domain,
by `feature-adaptation' of pre-trained
neural extractors using the semantic loss function constrained by abductive feedback.
\end{abstract}

\section{Introduction}

\label{sec;intro}
In ``Machine Intelligibility and the Duality Principle'' \cite{10.5555/646107.678977}, the
authors propose that a 2-way human interaction between human and 
machine necessarily brings up a `duality principle' in the construction
of software, defined as follows:

\begin{quote}
Software involved in human/computer inter-action should be designed at
two interconnected levels: a) a declarative, or self-aware level, supporting
ease of adaptation and human inter-action, and b) a procedural, or skill level,
supporting efficient and accurate computation.
\end{quote}

\noindent
Recent work in Machine Learning (ML) with highly accurate deep neural networks
(DNNs) has given rise to a similar kind of duality principle for constructing
DNN-based models with humans-in-the-loop. Communicating results from a DNN
clearly needs to be `human-friendly', referring to abstract concepts--like entities,
features, and relations--already known to the human. These
may not necessarily have any direct counterpart to the concepts actually being used
by the DNN to arrive at its prediction. Interest is also growing in
communicating human-knowledge to a DNN \cite{Dash2022ARO}, that is both
precise and reasonably natural. In \cite{10.5555/646107.678977}, symbolic logic is suggested
as a choice of representation for the declarative level of software concered
with human-interaction. The main reasons for this are that it is expressive enough
to capture complex concepts precisely, it
supports reasoning, and can be converted reasonably easily
to controlled forms of natural language (see for example \cite{schwitter-2010-controlled}).
The utility of a symbolic model for explanations  is reflected in substantial interest
in techniques such as LIME \cite{Ribeiro2016WhySI}, which results in an
embodiment of the duality principle by constructing human-intelligble models
for black-box predictors. In this paper, we are interested in the converse aspect
of the duality principle, namely: suppose we have a human-intelligible symbolic
theory, {\em a priori}, describing concepts and relationships that are expected
in intelligible explanations. Some situations where this can occur are:
(a) the theory may encode accepted knowledge in the domain, deliberately leaving
out data-specific aspects; 
(b) the theory  may have developed elsewhere, and we want to see if can
be adopted for local-use; or
(c) the theory may have been developed locally, but equipment changes may require
it to be `re-calibrated'.\footnote{Of course, there will be
cases where the symbolic theory itself may have to be revised, or even
completely re-constructed to suit local needs. We do not consider that here.}
Such a high-level declarative theory usually contains no mechanisms for linking
abstract concepts to actual raw data.
How should the lower (procedural) layer adapt to the provision of this high-level
specification? The authors in \cite{10.5555/646107.678977} do not offer any solutions.

In this paper, we adopt the position that the duality principle
entails a hierarchical system design, specified as a form of
function-decomposition. We examine a recent implementation \cite{Tsamoura2021NeuralSymbolicIA}
in which a DNN implements the low-level procedures
for linking high-level concepts in a domain-theory to raw data.
In itself, a hierarchical design for AI systems is not new, and has been adopted
at least from the early robotic applications like Shakey and Freddy, although not for
reasons of communicating with humans. But the increasingly widespread
use of DNNs as the preferred choice of ML models, and the increasing need for human
interaction requires us to confront issues of machine-intelligibility
arising from the use of neuro-symbolic systems with humans-in-the-loop.

\section{Hierarchical Neuro-Symbolic Modelling}
\label{sec:neurosym}
We assume the ML model implements a function $h:{\cal X} \mapsto {\cal Y}$. Here,
${\cal X}$ denotes the set of (raw) data instances, and ${\cal Y}$ denotes
the set of values of a dependent variable. Then
one formulation of $h$ as a hierarchical neuro-symbolic model
is to specify $h$ as the composition
of symbolic and neural functions. A specification for identifying models
of this kind is shown in Fig.~\ref{fig:spec}.


\begin{figure}
    \begin{description}
        \item[Given:] (a)  Data-instances $\{(x_i,y_i)\}_{i=1}^N$ of some
            task $T$, where $x_i \in {\cal X}$
            and $y_i \in {\cal Y}$;
        (b) A set of symbolic representations of the data ${\cal J}$; and
        (c) and a loss-function $L$
$L$;    \item[Find:] $n:{\cal X} \mapsto {\cal J}$
            and $s:{\cal J} \mapsto {\cal Y}$
            such that $\sum_i E[L(y_i,s(n(x_i))]$ is minimised.
    \end{description}
    \caption{A partial specification for hierarchical neuro-symbolic learning from
        data.}
    \label{fig:spec}
\end{figure}

Recently, an implementation for meeting the specification in Fig.~\ref{fig:spec}
with some restrictions has been proposed. We describe this next.

\subsection{$NEUROLOG$}
\label{sec:neurolog}

NEUROLOG \cite{Tsamoura2021NeuralSymbolicIA} is an implementation for learning
hierarchical neuro-symbolic models, with the following characteristics:
\begin{enumerate}
\item[(a)] The task $T$ is a  classification task, with the dependent
    variable $Y$ taking values from the set ${\cal Y}$, which has
    some distinguished class-label $\bot$;
\item[(b)] ${\cal J}$ is a set representing the data in a propositional logic.
        In ML-terms, this is entails representing
        the data by a finite set of features ${\cal F}$ = $\{f_1,f_2,\ldots,f_k\}$.
        For simplicity of exposition, we will take these features
        to be Boolean-valued, and ${\cal J} = \{0,1\}^k$;
\item[(c)] $s$ is known {\it a priori\/}. For each
    value $y \in {\cal Y}$, $s$ is assumed to be of the form:\footnote{
            Here,  ${\mathbf f}$ is
            a $k$-tuple of values assigned to $f_1,\ldots,f_k$ ($f=0$ denotes $f$ is $false$,
            and $f=1$ denotes $f$ is $true$); and
            $s_y({\mathbf f})$ = ($c_{1,y}|_{\mathbf f} \vee \allowbreak c_{2,y}|_{\mathbf f} \vee \allowbreak \cdots \vee \allowbreak c_{n_y,y}|_{\mathbf f}$).
            The $c_{i,y}$ are conjunctive formulae defined over values of features in
            ${\cal F}$.
}
\begin{equation*}
\begin{split}
s({\mathbf f})  & = y_1 ~~~~~~~{\mathrm{if}}~ s_{y_1}({\mathbf{f}})= true \\
                & =  y_2 ~~~~~~~{\mathrm{if}}~ s_{y_2}({\mathbf{f}}) = true \\
                & = \cdots \\
                & = \bot~~~~~~~{\mathrm{otherwise}}
\end{split}
\end{equation*}

\item[(d)] For each feature $f \in {\cal F}$, there
        are task-specific pre-processing functions $p_{T,f}:{\cal X} \mapsto X_f$. These
        are used to separate identify subsets of the raw data where the feature $f$
        takes specific values (like $0$ or $1$).
        Additionally, an overall task-specific function
        $p_T:{\cal X} \mapsto X_{f_1} \times X_{f_2} \times \cdots X_{f_k}$ is
        defined as $p_T(x) = (p_{f_1}(x),p_{f_2}(x),\cdots,p_{f_k}(x))$.
        The example below will make this clearer;
\item[(e)]  The neural function $n$ to be identified is now
        a composition of $p_T$ with
        $n': X_{f_1} \times X_{f_2} \times \cdots X_{f_k} \mapsto \{0,1\}^k$.
            That is, for $x \in {\cal X}$, 
            the prediction $Y = NEUROLOG(x)$, where
            $NEUROLOG(x) = s(n'(p_T(x)))$;
    \item[(f)] Given a training data-instance
            $(x_i,y_i)$ the implementation progressively updates parameters
            of the neural network implementing $n'$ by computing a
    `       semantic loss'$L(y_i,NEUROLOG(x_i))$ function that uses 
            {\em abductive feedback\/} from the conjuncts in $s_{y_i}$.
    \end{enumerate}

\begin{myexample}[$NEUROLOG$]
  In \cite{Tsamoura2021NeuralSymbolicIA}, the authors describe the implementation of
  $NEUROLOG$ using an example of a 3 $\times$ 3 chess-board with 1 black and 2 distinct white pieces. 
  The authors define a set of atomic predicates of the form $at(P,(X,Y))$ $X,Y \in \{1,2,3\}$. Each such grounded predicate indicates the presence
  of a chess piece at position (X,Y) where $P \in \{b(k), w(k), w(q), w(r), w(b), w(n), w(p)\} $.
  b(.) refers to a black piece, w(.) refers to a white piece and k,q,r,b,n,p refer to 
  king, queen, rook, bishop, knight and pawn respectively. Additionally, the authors provide a logical theory $T$ that 
  captures domain knowledge about the problem. The authors use a neural network to embed raw data into symbolic representations that can be used by the theory. The task specific preprocessing function $p_{T}$ involves segmenting each of the 9 blocks (features) on the 3 $\times$ 3 chess board into separate inputs. 
 \end{myexample}
   
\begin{figure*}[t!]
     \centering
     \subfigure[]{
         \includegraphics[width=13cm, height=5.2cm]{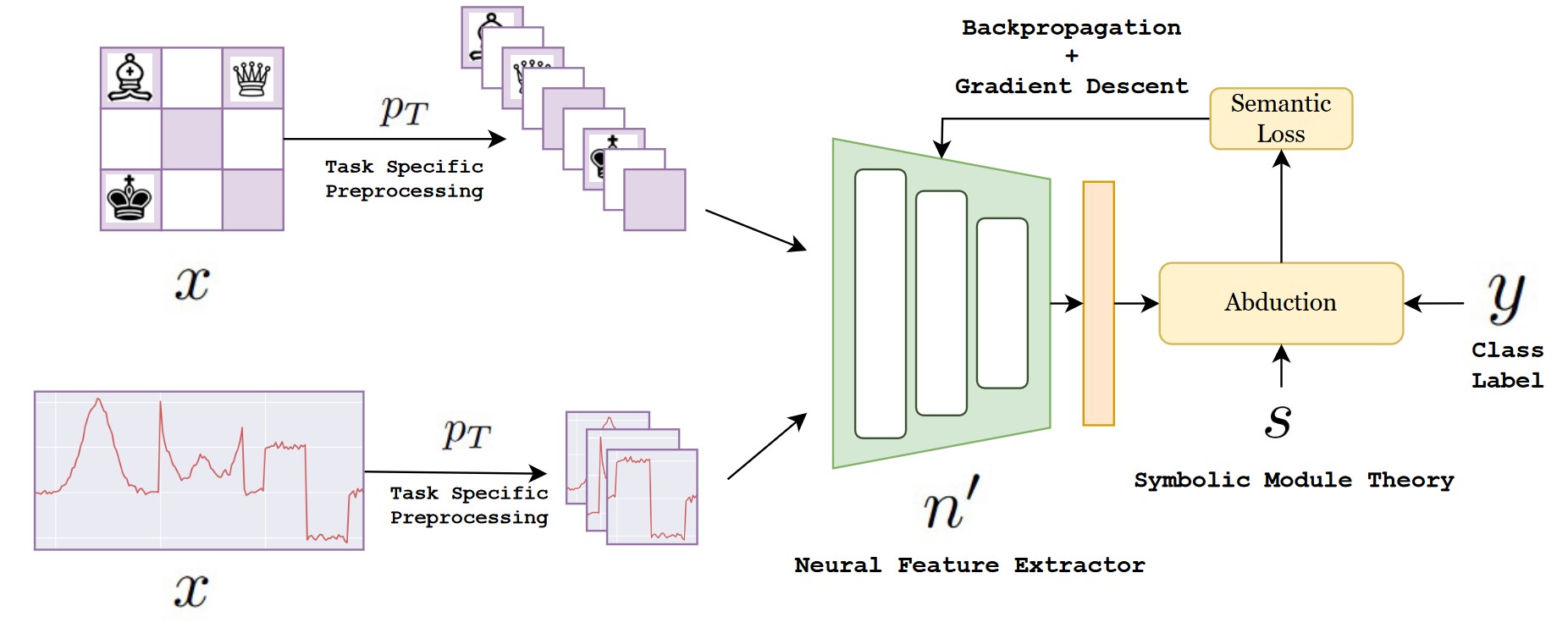}
         \label{fig:n}    
     }
     \subfigure[]{
         \includegraphics[width=10cm, height=5cm]{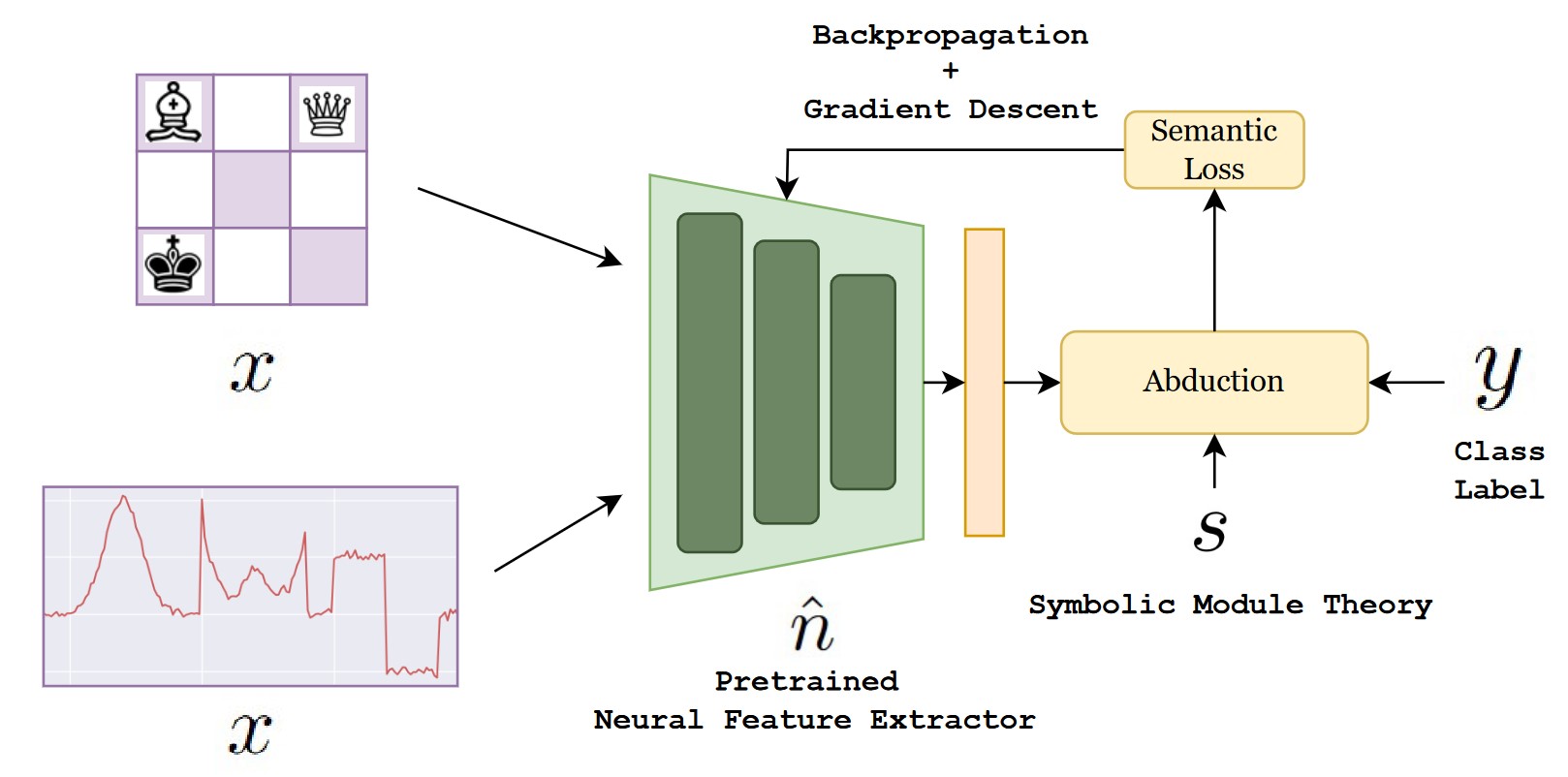}
         \label{fig:neurologminus}
     }
        \caption{(a) $NEUROLOG$: the image input is pre-processed by segmenting out individual squares, thereby making the downstream learning with semantic loss easier. The diagram is for the chess problem described in \cite{Tsamoura2021NeuralSymbolicIA}. (b) ${NEUROLOG}^-$: the image input is no longer pre-processed but the feature extraction network is potentially imperfect.}
\end{figure*}

    
  The task specific pre-processing could be seen as a form of prior
  information about the features, for realistic applications it may not always be
  obvious what these
  functions should be. We consider pre-training
  as a simple alternative for introducing
  prior information about the features.

\subsection{${NEUROLOG}$ without Pre-Processing}
\label{sec:neurologminus}

What is to be done if it is
not possible to specify or
implement the task-specific pre-processor
$p_T$? Since
the pre-processing step essentially constitutes
prior information about the task, we
can distinguish two kinds of implementations
of $n$: one that is obtained with no prior knowledge (correctly,
an uninformed prior);
and one that is obtained without task-specific pre-processing
but with task-independent prior information. For simplicity,
we will denote these two models as implementing
the functions $n_0$ (no prior) and $n_1$ (task-independent prior).
 The corresponding neuro-symbolic
model are $s(n_0(\cdot))$ and $s(n_1(\cdot))$,
which we denote by ${NEUROLOG}_0^-$ and
${NEUROLOG}_1^-$ (the minus superscript signifying
that it denotes ${NEUROLOG}$ without pre-processing).

One option for $n_1$ is to start with prior values for parameters
obtained from some form of pre-training.
That is, suppose there exists an implementation
$\hat{n}_1:\hat{{\cal X}} \mapsto \{0,1\}^k$, and the parameters
of this model are modified using the data drawn from ${\cal X}$ to
yield $n_1:{\cal X} \mapsto \{0,1\}^k$ that is
reasonably consistent (defined using the loss function) with the constraints 
imposed by $s$.
This may require ${\cal X} \subseteq \hat{{\cal X}}$, or
at least a sufficient
large overlap between ${\cal X}$ and $\hat{{\cal X}}$).
Further, we would expect the neuro-symbolic model
obtained in this way not to be as effective as
one with task-specific pre-processing.
But, it allows the practically useful prospect of implementing a form of
$NEUROLOG$ of `feature-adaptation' 
(see Fig.~\ref{fig:neurologminus}).
The extent to which this can help when
pre-processing is not possible
is the focus of the experimental investigation below.



\section{Experimental Evaluation}
\label{sec:expt}

\subsection{Aims}
\label{sec:aims}

For brevity, we use $N(\cdot)$ to denote $NEUROLOG$ with
pre-processing; $N_0^-$ to denote ${NEUROLOG}_0^-$ (that is,
$NEUROLOG$ without pre-processing and without a pre-trained feature-extractor);
and $N_1^-$ to denote ${NEUROLOG}_1^-$ (that is,
$NEUROLOG$ without pre-processing and a with a pre-trained feature-extractors).
We aim to investigate the following conjectures:

\begin{description}
    \item[Pre-Processing.] If pre-processing is possible, then models constructed by $N$
        are better than those constructed by $N_0^-$ and $N_1^-$;
    \item[Pre-Training.] If pre-processing is not possible, then models constructed
        by $N_1^-$ are better than models constructed by $N_0^-$.
\end{description}

The following clarifications are necessary: (a) By ``better'', we will mean higher
predictive accuracy and higher explanatory fidelity (a definition for this is in
the Methods section); and (b) Using synthetic problems and simulated data, we are
able to obtain $N_1^-$ models starting from progressively poorer pre-trained
models. This allows us to examine qualifications to the conjectures,
based on the corresponding $N_1^-$ models.


\subsection{Materials}
\label{sec:mat}

\subsubsection{Problems}

We report results from experiments conducted on two different problem domains: the Chess problem reported in \cite{Tsamoura2021NeuralSymbolicIA}
and synthetic time-series data obtained by controlled simulation. \\

\noindent
\textbf{Chess.} We refer the reader to \cite{Tsamoura2021NeuralSymbolicIA} for
    details of this problem and simply summarise the main
    aspects here. The raw data consists of a 3 $\times$ 3 chess board with 3 chess pieces -- a black king and two distinct white pieces. Task-specific pre-processing involves a `segmentation' step that separates each data-instance
    into the 9 squares in a pre-specified order. Each
    square corresponds to a (multi-valued) feature, with values indicating whether it is empty, or which of 7 other
    pieces it has (black king, white king, white rook,
    white knight, white pawn, white bishop, white knight). The logical theory encodes
    the conditions for class-labels
    denoting: 
    safe, draw, and mate according to the usual rules of chess (or `illegal' otherwise). In the
    terminology of this paper, this corresponds to $m$
Yes.    definitions for each class-label (the number $m$ is different for each class) in terms of a
    disjunct of conjunctions of the 9 features (with $\bot$ = `illegal').



\begin{figure*}[t!]
     \centering
     \subfigure[]{
         \includegraphics[width=8cm, height=2cm]{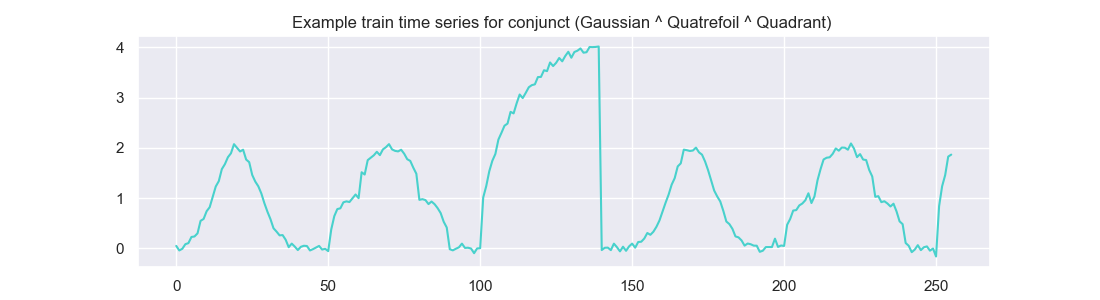}
         \label{f2a}
     }
     \subfigure[]{
         \includegraphics[width=8cm, height=2cm]{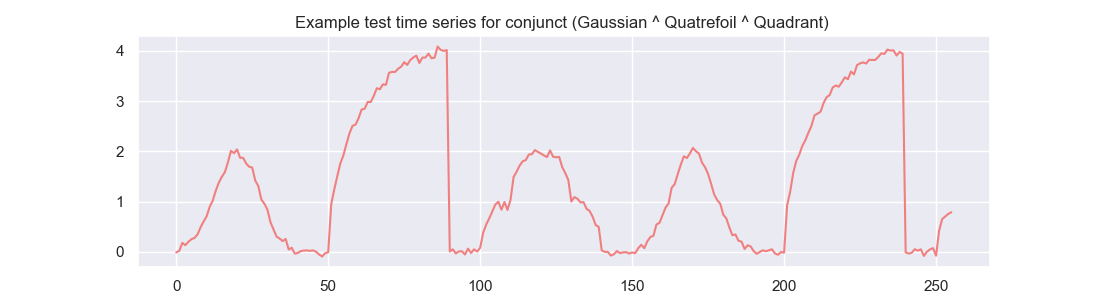}
         \label{f2b}
     } \\
     \subfigure[]{
         \includegraphics[width=8cm, height=2cm]{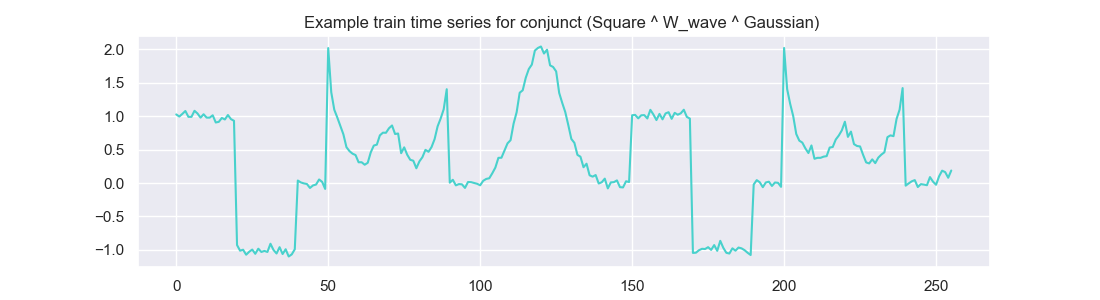}
         \label{f2c}
     } 
     \subfigure[]{
         \includegraphics[width=8cm, height=2cm]{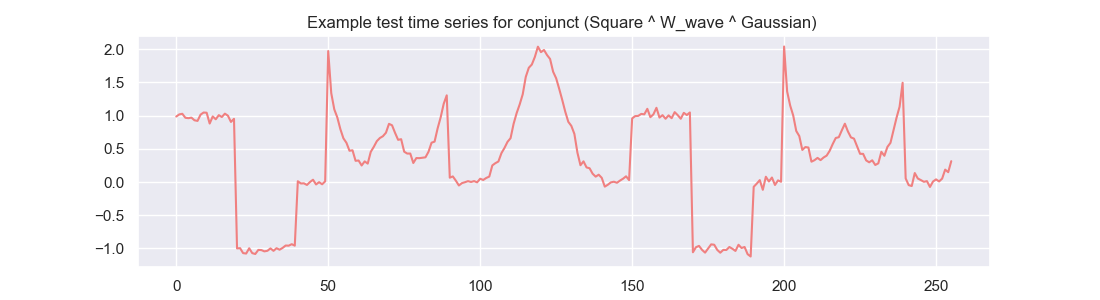}
         \label{f2d}
     }
        \caption{(a, b) Examples of randomly generated train and test time-series data for the conjunct (\textit{Gaussian} $\land$ \textit{Quatrefoil} $\land$ \textit{Quadrant}). (c, d) Examples of randomly generated train and test time-series data for the conjunct (\textit{SquareWave} $\land$ \textit{W\_wave} $\land$ \textit{Gaussian}). Note that the order in which the shapes occur can be different between a train and test data point, as in (c, d).}
        \label{fig:timex}
\end{figure*}

    We use the same data provided with and code adapted from the author's implementation\footnote{https://bitbucket.org/tsamoura/neurolog/src/master/} of $NEUROLOG$ for our experiments.
\\
\noindent
{\bf Time Series.} We generate synthetic
examples of simple time-series data, to resemble
aspects of the Chess problem. 
Each time-series consists of three features in the form of different shapes, sampled from a pool of nine shapes:
\{\textit{Blank, SemiCircle, SquareWave, Triangle, Gaussian, Quadrant, Trapezium, Quatrefoil, W-wave}\}. Each shape
spans 50 time steps and three of the these shapes are
sampled and concatenated end-to-end to form one cycle. One
time series is a periodic repetition of approximately 1.5
such cycles, extending over 256 time-steps. An example
of some randomly generated time-series is shown in
Fig.~\ref{fig:timex}.

\noindent
As with the Chess data, it is possible to devise a
`segmented' input for $NEUROLOG$ that converts a time-series
into 3 features in a pre-defined order (left-to-right; $f_1$,
$f_2$, $f_3$). A symbolic `theory' about the data
is constructed as follows, by assuming a set of
4 class-labels (A,B,C, and D). A definition of
a class-label essentially consists of a random number
of conjuncts, each with 3 randomly chosen shapes
(the details are in ``Methods''). Data for each class
label are then generated to be consistent with the
target theory for that class label.

\subsubsection{Machines and Environments}
All the experiments were run on machines with Intel(R) Xeon(R) E5-2698 v4 CPU end to end with an allocation of 8GB of memory. We use PyTorch to program the Neural Networks. $NEUROLOG$ uses A-System \cite{Nuffelen2001AsystemDP} running over SICStus Prolog 4.5.1 for the abduction process. However, for our experiments, we cache the abductive feedback for all the three classes in the chess domain in text files and use those for calculating the semantic loss.
 
\subsection{Methods}
\label{sec:method}

We first describe some procedures required for conducting the experiments.

\subsubsection*{Generation of Time-Series Problems}

Recall the time-series problems consist
of a periodically repeating pattern of 3 shapes drawn from
a possible set of 9 shapes. For the experiments, we need,
a symbolic theory and data instances
(time-series) consistent with the symbolic theory. For
our experiments, the symbolic theory $s$ will assign
one of 4 class-labels to data instances. The definition
used by $s$ for each class-label is obtained as follows:

\begin{enumerate}
    \item We partition the set of nine shapes into 
        groups of three features each. These groups are
        assigned to one of the three features each.  This is
        done randomly. Each feature can only take one of 
        the shapes from the assigned group at a time as a possible value.
        \item Next, we randomly sample one shape from the three
        groups each. The combination of these three shapes forms a randomly sampled conjunct.
    \item Next, we decide the number of conjuncts to be
        assigned to each class. We define the parameters
        \textit{lower bound} and \textit{upper bound}. For each
        class in the theory, number of conjuncts to be assigned
        is determined by randomly sampling a number from
        [\textit{lower bound}, \textit{upper bound}). Note that
        different classes can have different number of
        conjuncts.
    \item For each class, we randomly sample the conjuncts
        required as discussed in Step 1. We take care that
        conjuncts do not overlap between classes.
    
\end{enumerate}

An example of a randomly sampled $s$ is given below:

\noindent
The result of these steps is a randomly drawn symbolic
theory about the data-instances. Using this theory,
data-instances consistent with the theory are generated
as follows:

\begin{enumerate}
    \item For each conjunct assigned to a particular class,
    we generate a cycle by concatenating the shapes in the conjunct. Let's denote these three shapes as $\psi_1$, $\psi_2$ and $\psi_3$.
    \item To differentiate the test data from the train data, we concatenate the shapes in the order $\psi_1, \psi_2, \psi_3$ for the training data, whereas for test data, 30\% of the time series are constructed by concatenating them in the same order and the rest are constructed by concatenating them them in the order $\psi_1, \psi_3, \psi_2$ (see Fig.~\ref{fig:timex}).
    \item The cycle is repeated to get a total of 10000 time steps for the training data and 750 time steps for testing data.
    \item We add Gaussian noise to these time series to add randomness.
    \item These time series are then cut into non-overlapping sequences of length 256 each giving us a total of 39 samples per conjunct for training data and 9 samples for testing data. Note that the total amount of training and test data for the time-series experiments depends on the total number of conjuncts in the theory, which varies across random runs.
\end{enumerate}


\subsubsection*{Task-independent Pre-training}

The steps followed for testing the conjectures
is straightforward:
is:
    \begin{itemize}
        \item[] Repeat $R$ times:
        \begin{enumerate}
            \item For $T=$ $Chess$, $Time-Series$
            \begin{enumerate}
                \item Randomly generate training and test data as described above (applicable only for time series data)
                \item Let $s_T$ be the symbolic theory
                    for task $T$
                \item Let $p_T$ be the task-specific
                    pre-processor for task $T$ and
                    $N$ denote the $NEUROLOG$ model
                    obtained using the training data, along
                    with $s_T$ and the pre-processor $p_T$
                \item Let $\hat{n}_\alpha$ be an approximate
                    pre-trained model for the features ($\alpha$ denotes
                    an approximation parameter: see details below) and $N_{1,\alpha}^-$
                    be the ${NEUROLOG}^-$ model obtained using the training
                    data, along with $s_T$ and with parameters of the neural
                    feature-extractor initialised using the parameters
                    of $\hat{n}_\alpha$
                \item Let $N_0^-$ be the ${NEUROLOG}^-$ model obtained
                    using the training data, along with $s_T$ and
                    with parameters of a completely untrained
                    neural feature-extractor;
                \item Record estimates of predictive accuracy and explanatory fidelity of $N$, $N_{1,\alpha}^-$, and $N_0^-$ on
                    the test data (see details below)
            \end{enumerate}
            \item Test the ``Pre-Processing'' conjecture by
                    comparing the (mean) estimates of predictive accuracy and
                    explanatory fidelity for $N$ against those obtained
                    for $N_{1,\cdot}^-$ and $N_0^-$
            \item Test the ``Pre-Training'' conjecture by comparing the
                (mean) estimates of predictive accuracy and explanatory
                fidelity for $N_{1,\cdot}^-$  againsts those obtained
                for $N_0^-$.
        \end{enumerate}
    \end{itemize}

\noindent
The following details are relevant:

\begin{itemize}
    \item For all experiments $R = 5$;
    \item Training and test data sizes for Chess are 9000 and 2000 respectively.
        For Time Series the corresponding numbers are about 1000 and 200
        (numbers vary due to the theory being sampled randomly); We use \textit{lower bound} = 5 and \textit{upper bound} = 8.
    \item We restrict $\alpha$ to $0.1$, $0.2$ and $0.3$; denoting
        low, medium and high levels of difference with the true
        feature detector
    \item Predictive accuracy is the usual ratio of correct predictions
        to total instances predicted. The estimates are from predictions on the test data
        instances.
    \item Explanatory fidelity refers to the ratio of correctly explained instances
        to the total number of instances. An instance is correctly explained
        if the conjunct used to generate the class label is the (only) conjunct
        that is determined to be true given the features extracted by the neural layer;
    \item We report the metric after taking a running average to 100.
    \item Model performance is represented by the pair $(P,E)$, where
        $P$ denotes predictive accuracy and $E$ denotes explanatory
        fidelity. Let model $M1$ have performance $(P1,E1)$ and
        model $M2$ have performance $(P2,E2)$.
        If $P1 > P2$ and $E1 > E2$ then we will say model $M1$ is
        better than model $M2$.
\end{itemize}

\begin{table*}[t!]
\parbox{.45\linewidth}{
\centering
\begin{tabular}{|cc|c|c|}
\hline
                                  &                   & \textbf{Pred Acc.(\%)} & \textbf{Expl Fid.(\%)}\\
\hline
$N$                                &                   &         95.41 (0.66)             &         94.67 (0.69)              \\
\hline \hline
\multirow{3}{*}{$N_{1,\alpha}^{-}$}    &   $\alpha = 0.1$  &        84.71 (1.34)               &         62.71 (3.80)              \\

                                  &   $\alpha = 0.2$  &        80.31 (1.67)               &         48.12 (5.07)           \\

                                  &   $\alpha = 0.3$  &        78.79 (2.68)              &         43.32 (4.17)             \\
\hline
$N_0^{-}$                              &                   &         79.88 (1.57)              &         0.10 (0.13)              \\
\hline
\end{tabular}
\label{chess}
\begin{center}
(a) Chess
\end{center}
}
\hfill
\parbox{.45\linewidth}{
\centering
\begin{tabular}{|cc|c|c|}
\hline
                                  &                   & \textbf{Pred Acc.(\%)}           & \textbf{Expl Fid.(\%)}\\
\hline
$N$                                &                   &         100 (0.00)                &    100 (0.00)                          \\
\hline \hline
\multirow{3}{*}{$N_{1,\alpha}^{-}$}&   $\alpha = 0.1$  &         82.07 (6.59)          &        76.05 (10.23)                       \\

                                  &   $\alpha = 0.2$  &         81.48 (4.44)          &        76.08 (7.78)                       \\

                                  &   $\alpha = 0.3$  &         81.56 (4.61)              &        75.37 (10.28)                       \\
                                   


\hline
$N_0^{-}$                          &                   &         75.79 (11.86)                &      36.81 (16.85)                   \\
\hline
\end{tabular}
\label{ts}
\begin{center}
    (b) Time Series
\end{center}
}
\caption{Performance of models with pre-processsing ($N$); without pre-processing
    but with pre-training ($N_{1,\alpha}^-$); without pre-processing and without
    pre-training ($N_{0}^-$. ``Pred. Acc.'' refers to (mean) predictive accuracy;
    ``Expl. Fid.'' refers to (mean) explanatory fidelity. The number in brackets is
    the standard deviation. Lower values of $\alpha$ indicates that the pre-trained
    approximation $\hat{n}_\alpha$ is closer to the correct model for feature-subset
    identification (the values of $\alpha$ correspond to
    low, medium, and high-levels of difference between predictions from $\hat{n}$ and the correct prediction.}
\label{tab:results}
\end{table*}

\subsubsection*{Approximate Pre-Training ($\alpha$)}

Approximate models for pre-training are generated by deliberately altering
feature-labels.
The value of $\alpha$ defines the percentage of feature labels that are corrupted during the time of pretraining.
\begin{itemize}
    \item For the chess experiment, the position and identity of each of the three pieces has a probability of $\alpha$ to be changed to one of the other possible values which is randomly chosen.
    \item For the time series experiment, the labels for each of the three shapes has a probability of $\alpha$ of being changed to one of the other two possible values (i.e. the remaining two values from the same "group").
\end{itemize}

\subsection{Results}
\label{sec:results}

\begin{figure*}[t!]
     \centering
     \subfigure[]{
         \includegraphics[width=8cm, height=5cm]{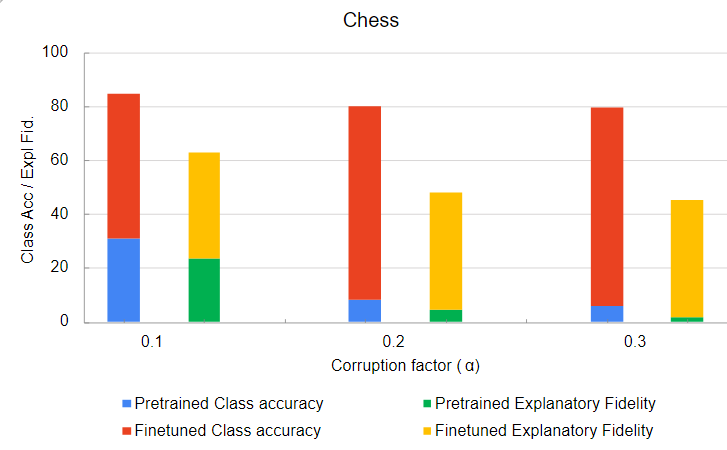}
         \label{ba}
     }
     \subfigure[]{
         \includegraphics[width=8cm, height=5cm]{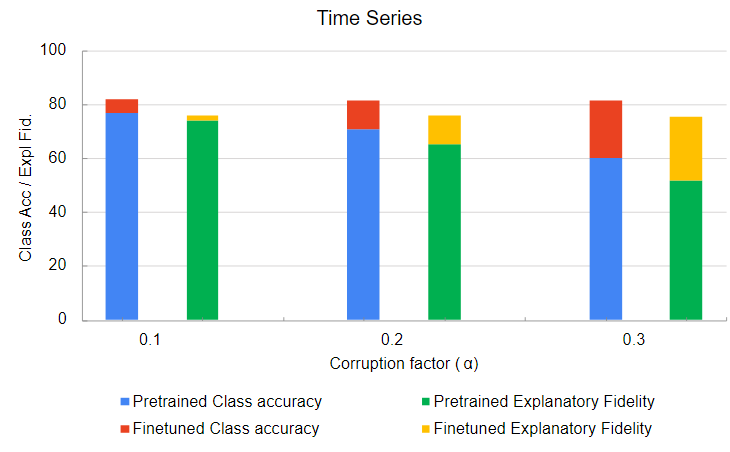}
         \label{bb}
     } 
    \caption{Increase in Prediction Accuracy and Explanatory Fidelity of the approximate models after finetuning with semantic loss}
    \label{bars}
\end{figure*}

The main results of the empirical study are tabulated in
Tables \ref{tab:results}(a), (b). The
tabulations suggest the following:
(a) If pre-processing is possible, then clearly
    better models (that is, with higher predictive accuracy and higher explanatory fidelity)
    result from using task-specific
    pre-processing than from using task-independent pre-training.
(b) If pre-processing is not possible, then
    it is usually better to employ a 
    pre-trained model than to start with
    an uninformed model;
(c) As expected, as the pre-trained approximation  becomes
    progressively uninformative (medium to high values of $\alpha$), performance on predictive and explanatory
    fronts decreases. However, even with with fairly
    poor initial approximation, explanatory fidelity
    remains substantially better than
    starting with an uninformed model; and
(d) The loss in performance due to
    lack of pre-processing can be offset to some extent
    through the use of pre-training, especially with
    a good initial approximation (low value of $\alpha$).
Taken together, these results provide empirical
support for the
Pre-Processing and Pre-Training
conjectures.

\noindent 
We now turn to a more detailed finding from the experiments
for cases when pre-processing is not an option. It
is evident from Table \ref{tab:results} that
there is a substantial `gap' between the predictive
accuracy and explantory fidelity estimate.
This is more pronounced as the pre-trained
approximator  gets worse; and especially obvious
for the model $N_0^-$ that does not employ a pre-trained approximator.
        That is, there are more predictions that are
        correct, but for the wrong reasons. We find
        that this is because without sufficient prior
        information about the features, the neural networks
        training process allows a convergence to 
        arbitrary conjuncts that are associated with
        the correct class label. This tends to keep
        predictive accuracy from falling, but results
        in substantially poorer accuracy of correct
        feature-identification. This is shown
        for Chess in Table \ref{chesssegnosegfeat}, where the effect is especially  pronounced (the
        results from the model with pre-processing are shown
        purely for reference). We also show the comparision between the approximate model $n_{\alpha}$ and $N_{1,\alpha}^{-}$ in Fig. \ref{bars}. The increase in explanatory fidelity is clearly visible and is more significant in the case of chess domain.
\begin{table*}[t]
\centering

\begin{tabular}{|cc|c|c|c|c|c|c|c|c|c|}
\hline
Model &  &\multicolumn{9}{|c|}{Mean F-score} \\
\hline
 & & $f_1$ & $f_2$  & $f_3$ & $f_4$ & $f_5$ & $f_6$ & $f_7$ & $f_8$ & $f_9$  \\
\hline
\multirow{3}{*}{$N_{1,\alpha}^{-}$} & $\alpha = 0.1$ & 0.94 & 0.95 & 0.91 & 0.96 & 0.98 & 0.95 & 0.95 & 0.95 & 0.94 \\

                                    & $\alpha = 0.2$ & 0.94 & 0.93 & 0.89 & 0.94 & 0.97 & 0.91 & 0.91 & 0.93 & 0.92 \\

                                    & $\alpha = 0.3$ & 0.91 & 0.94 & 0.85 & 0.92 & 0.89 & 0.91 & 0.88 & 0.86 & 0.92 \\
\hline
$N_0^{-}$ &  & 0.22 & 0.13 & 0.29 & 0.50 & 0.58 & 0.30 & 0.30 & 0.25 & 0.36 \\
\hline \hline
$N$ & & 0.99 & 0.99 & 0.99 & 0.99 & 0.99 & 0.99 & 0.99 & 0.99 & 0.99
 \\
\hline
\end{tabular}
\caption{Comparing mean F1-scores for feature-prediction
    in the chess domain.
    The models are as in Table \ref{tab:results}.}
\label{chesssegnosegfeat}
\end{table*}

    The symbolic theory for Chess is significantly more complex than
    the ones used for the Time Series data (the conjunct size in
    the Chess theory is 9, compared to 3 for Time Series and the number of conjuncts belonging to each class in Chess is also significantly higher). We conjecture
    that if pre-processing is not possible, then for complex theories
    pre-training to some extent may be needed to attain reasonable levels of explanatory fidelity.

\section{Related Work}
While the impressive performance of deep learning architectures underscores the pattern recognition capabilities of neural networks, deep models still struggle to imbibe explicit domain knowledge and perform logical reasoning. Symbolic systems on the other hand are adept at reasoning over explicit knowledge but can only ingest symbolic inputs. The community has been striving to combine the pattern recognition capabilities of neural networks with the reasoning and knowledge representation  power of symbolic techniques. Towards this end, one line of attack has been to feed knowledge rich symbolic features explicitly as inputs to a neural network instead of relying on the network to learn these features from low level data \citep{Lodhi,ashwin_ILP,ashwin_MLJ}. These techniques are useful when i) Knowledge regarding high level symbolic features is easily accessible and ii) The high level features are easily computed from low level data. 

The second line of attack involves using a neural network to ingest noisy unstructured real world data (Images, Time Series, Speech or Text) and predict the symbolic features that can be ingested by a symbolic model \cite{deepreader}. In many situations, although knowledge about the appropriate symbolic features is available and a symbolic theory for making predictions exists, the raw input data is not easily transformed into the symbolic inputs necessary for the symbolic theory. Doing so accurately via neural networks would require a large volume of annotated data for each symbolic input which is often infeasible to obtain for a new domain. Recent efforts towards end-to-end neuro symbolic training \citep{deepproblog, neurolog} aim to address this limitation by obviating the need to learn neural feature detectors in isolation prior to integration with the symbolic theory. This paper is concerned with this type of neuro symbolic architecture.

Among systems that attempt to replace symbolic computations with differentiable functions 
\cite{gaunt} develop  a framework for creation of end-to-end trainable systems that learn to write interpretable algorithms with perceptual components, however the transformation of the theory into differentiable functions is restricted to work for a subset of possible theories. Logic Tensor Networks (LTNs)\cite{donadello} integrate neural networks with first-order fuzzy logic to allow for logically constrained learning from noisy data. Along similar lines the DeepProbLog\cite{deepproblog} system introduces the notion of neural predicates to probabilistic logic programming\cite{problog} that allows for backpropogation across both the neural and symbolic modules. The ABL  \cite{dai} system was the first to use abductive reasoning to refine neural predictions via consistency checking between the predicted neural features and the theory. This system was recently refined by using a similarity based  consistency optimization \cite{huang}, that relies on assumptions about the inter class and intra class neural features. Recently  Abductive Meta-Interpretive Learning\cite{daisteve}  was proposed to jointly learn  neural feature predictors and the logical theory via induction and abduction.
While the above abduction based approaches require iterative retraining of the neural module, and only utilize some of the abduced inputs for backpropagation, the NEUROLOG system \cite{Tsamoura2021NeuralSymbolicIA} uses the complete set of abduced inputs and backpropogates errors using Semantic loss \cite{xu} that more fully capture theory semantics. 

\section{Concluding Remarks}
A limitation of  prior abduction based neuro-symbolic approaches is that they only experiment with cleanly delineated inputs and the neural network does not have the additional burden of learning where to attend while predicting a feature. However, in many situations such feature delineation is not possible. Consider a dataset of images containing multiple objects at random locations and a logical feature corresponding to whether the number of cars in an image is greater than a threshold. In this case it is not obvious how one would slice the image, and therefore the burden falls on the neural network to extract this feature directly from the full image. We demonstrate with NEUROLOG that for such cases where the network is forced to operate on the full inputs, and is trained only with semantic loss, it fails to learn the correct sub-symbolic features but can still yield high predictive accuracy. We further show that the resulting model can be improved to a degree with even partially pre-trained neural models with the extent of the improvement dependent on the quality of the initial pre-training. This is very useful for situations where the same theory applies to multiple domains with varying distributions of sub-symbolic features. The imperfect feature detectors trained on a different distribution can be significantly improved upon with semantic loss and abductive feedback. 
The results from our experiments reinforce the message: {\em Pre-process. Otherwise pre-train\/}.

\label{sec:conc}
\bibliography{aaai23}

\clearpage

\onecolumn
\section{Appendix}

\section{Data}

\subsection{Using MNIST Digits for Chess Data}
\begin{figure}[h]
\centering
\includegraphics[width=4cm, height=4cm]{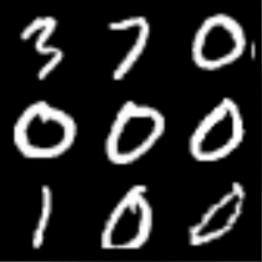}
\caption{Image used for a chess board having a \textit{White Bishop} at $(1,1)$,
a \textit{White Queen} at $(1,2)$ and a \textit{Black King} at $(3,1)$}
\label{fig:chess_board}
\end{figure}

As done in the original NEUROLOG
experiments\footnote{https://bitbucket.org/tsamoura/neurolog/src/master/} we use
the MNIST\footnote{http://yann.lecun.com/exdb/mnist/} handwritten digits dataset
for performing experiments on the chess domain. Each of the eight possible
values for a position on the 3 $\times$ 3 chess board is mapped to a number from
0 - 7 as given: \textit{Empty} - 0, \textit{Black King} - 1, \textit{White Rook}
- 2, \textit{White Bishop} - 3, \textit{White Knight} - 4, \textit{White King} -
5, \textit{White Pawn} - 6, \textit{White Queen} - 7. An image of a chess board
is made by placing an instance of an image of a digit from the MNIST dataset
corresponding to a chess piece (or \textit{Empty}) at it's respective position.
Each MNIST image has a dimension of 28 $\times$ 28. Hence, an image of a chess
board has the dimensions 84 $\times$ 84. Fig. \ref{fig:chess_board} shows an
example image of a chess board in a given configuration.

\subsection{Logical Theory Used for Time Series Experiments}
Methods section in the main paper describes how we randomly sample logical
theories for experiments on time-series data. Table \ref{table:eg_theory} shows
one such example theory where the \textbf{Class} column contains the four
classes that we use for time-series domain experiments and the
\textbf{Conjuncts} columns contains disjunction of conjuncts assigned to a given
class. This theory was sampled using \textit{lower bound} = 2 and \textit{upperbound} = 5. For brevity. we use numbers to represent shapes: 0 - \textit{Blank},
1 - \textit{SemiCircle}, 2 - \textit{Triangle}, 3 - \textit{Gaussian}, 4 -
\textit{SquareWave}, 5 - \textit{Quadrant}, 6 - \textit{Trapezium}, 7 -
\textit{Quatrefoil}, 8 - \textit{W-wave}.

\begin{table}[h]
\centering
\begin{tabular}{|c|c|}
\hline
\textbf{Class}                        & \textbf{Conjuncts} \\ 
\hline
A   &  $(2 \land 0 \land 7 ) \lor (4 \land 3 \land 6) \lor (1 \land 5 \land 8)$         \\
\hline
B   &  $(4 \land 5 \land 8) \lor (1 \land 0 \land 7)$       \\
\hline
C   &  $(2 \land 5 \land 6) \lor (1 \land 3 \land 7) \lor (4 \land 0 \land 8)$           \\
\hline
D   &  $(1 \land 5 \land 7)  \lor (4 \land 1 \land 6)$  \\
\hline
\end{tabular}
\caption{Example Theory for an experiment on Time-Series Data}
\label{table:eg_theory}
\end{table}

\subsection{Shapes used for Time-Series Experiments}

\begin{figure}[h]
     \centering
     \subfigure[Blank]{
         \includegraphics[scale = 0.2]{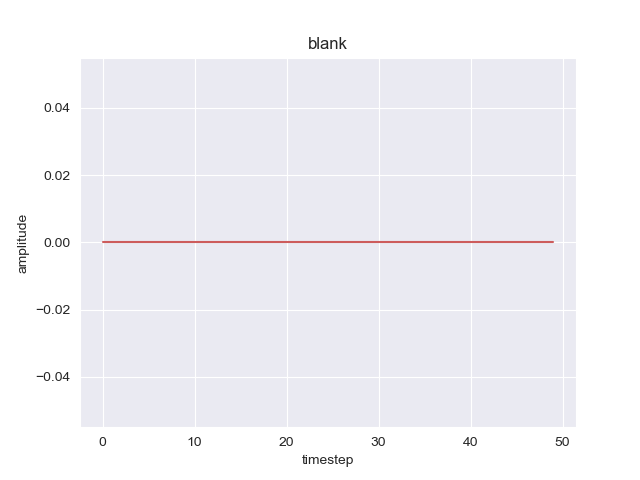}
     }
     \subfigure[SemiCircle]{
         \includegraphics[scale = 0.2]{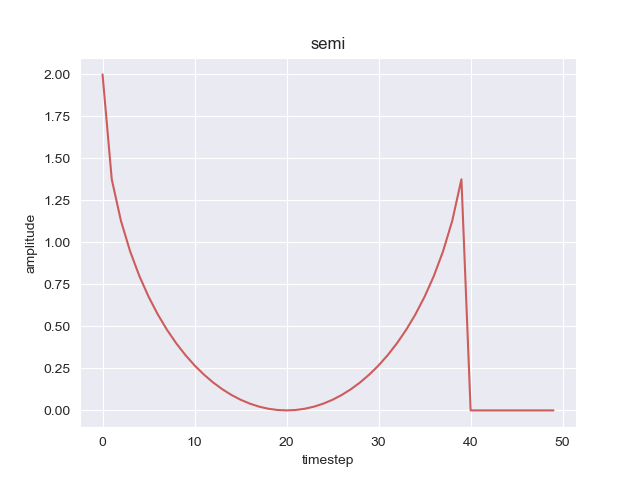}
     }
     \subfigure[Triangle]{
         \includegraphics[scale = 0.2]{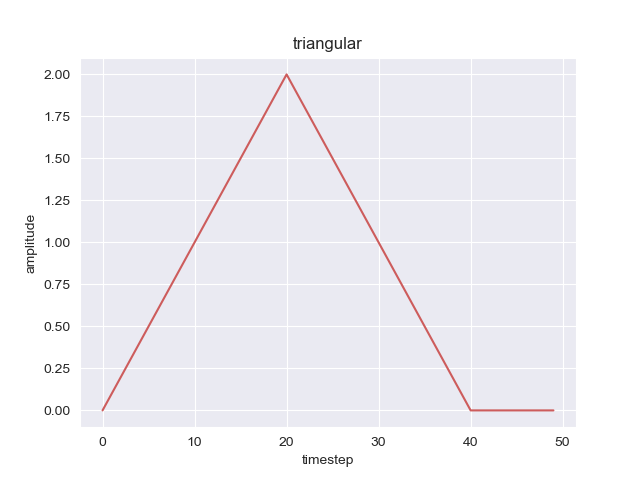}
     } \\
     \subfigure[Gaussian]{
         \includegraphics[scale = 0.2]{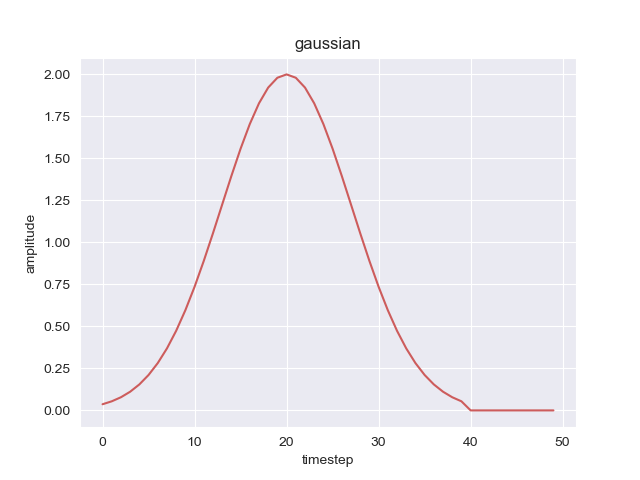}
     }
     \subfigure[SquareWave]{
         \includegraphics[scale = 0.2]{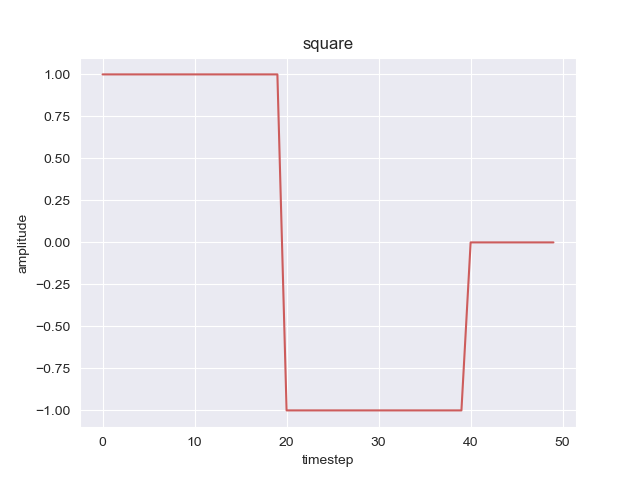}
     }
     \subfigure[Quadrant]{
         \includegraphics[scale = 0.2]{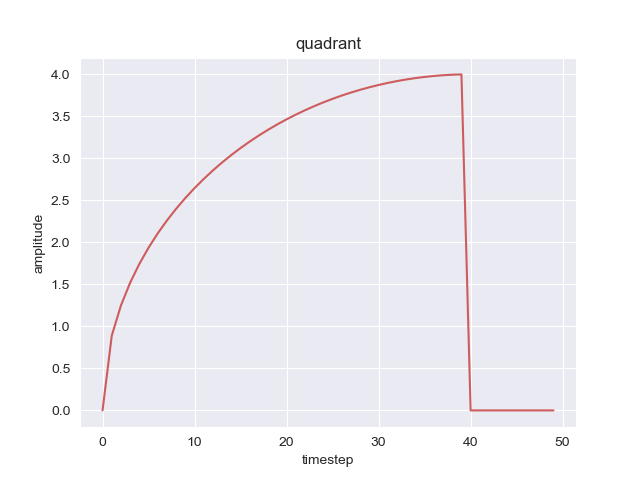}
     } \\
     \subfigure[Trapezium]{
         \includegraphics[scale = 0.2]{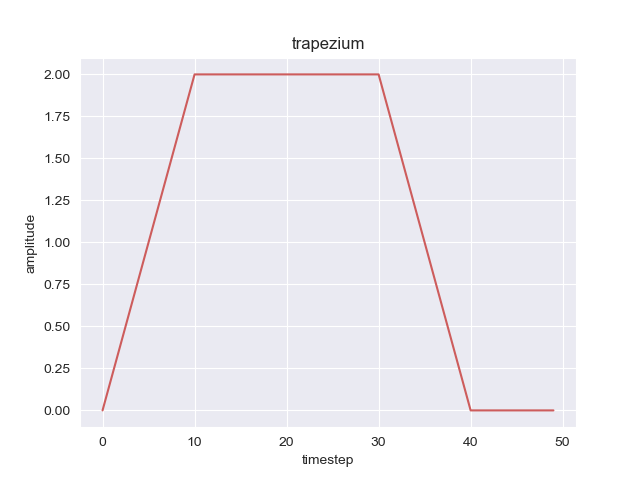}
     }
     \subfigure[Quatrefoil]{
         \includegraphics[scale = 0.2]{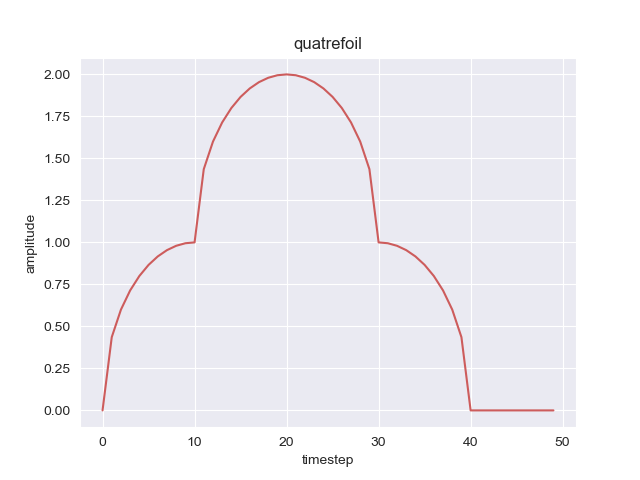}
     }
     \subfigure[W-wave]{
         \includegraphics[scale = 0.2]{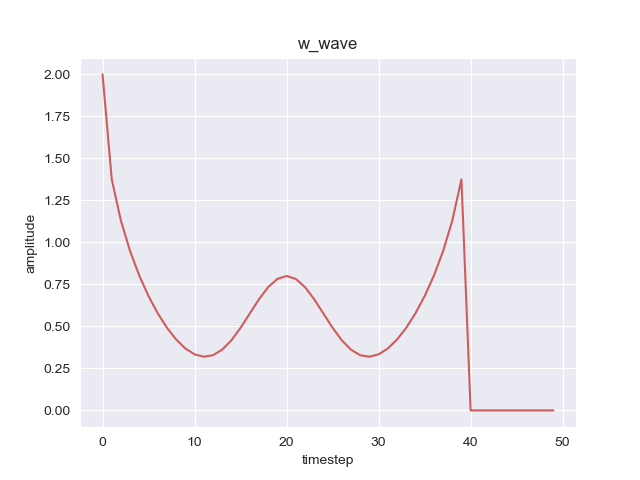}
     } \\
        \caption{Figures and equations of 9 shapes used for making the synthetic time-series data}
        \label{fig:shapes}
\end{figure}

The data used for time-series experiments is made by composing shapes derived
from a set of 9 shapes which are generated procedurally. Each shape spans 50
time-steps. Fig. \ref{fig:shapes} shows all of the 9 shapes and the equations
used to generate them.
3 of such shapes are joined together and a guassian noise of mean 0 and standard
deviation of 0.05 is added to make the timeseries.

\subsection{Segmenting Features for NEUROLOG Experiments on Time-Series Data}
Running $NEUROLOG$ on the time-series domain requires a neat segmentation of the
three features (shapes) present in a time-series. Each shape spans 50
time-steps. We use this knowledge to perform the segmentation. For $NEUROLOG$
experiments, we split the training and test series in lengths of 150 instead of
256. This is to ensure that the start of each time-series data point coincides
with the start of a shape. Then we break down the 150 time-steps into 3
consecutive segments of 50 time-steps each which correspond to the three
features.

\section{Corrupting Feature Labels}

\subsection{Corrupting Piece Information in Chess Data}

Let Let $\mathcal P$ be the set of all chess pieces in the chess domain $\cal C$ be a set defined as below and $\alpha$ with is the corruption factor

\begin{center}
    $\mathcal{P} = \{b(k), w(r), w(b), w(n), w(k), w(p), w(q)\}$ \\
    $\mathcal{C} = \{1, 2, 3\}$
\end{center}

In the chess domain, a feature label for a particular instance of a chess board
can be represented as given below where where $P_i \in \mathcal{P}$ are the
identities of the chess piece and $X_i,Y_i \in \mathcal{C}$ are their
coordinates on the chess board, with the constraint that one of the pieces is a
black king and the other two pieces are distinct white pieces.

\begin{center}
    $\{at(P_1,(X_1,Y_1)), at(P_2,(X_2,Y_2)), at(P_3,(X_3,Y_3))\}$
\end{center}

To corrupt the feature label for a particular instance of chess data, the steps given below are executed, each with a probability of $\alpha$.

\begin{center}
    $P_i \leftarrow p;  p \sim \mathcal{P} \slash \{P_i\}$  $\forall i \in \{1,2,3\}$ \\
    $X_i \leftarrow x;  x \sim \mathcal{C} \slash \{X_i\}$  $\forall i \in \{1,2,3\}$ \\
    $Y_i \leftarrow y;  y \sim \mathcal{P} \slash \{Y_i\}$  $\forall i \in \{1,2,3\}$ \\
\end{center}

\subsection{Corrupting Shape labels for Time-Series Data}

As described in the main paper, the set of 9 shapes is randomly partitioned into
a 3 sets of 3 shapes and each of these sets is assigned to the three features of
the time-series. Let $\mathcal{S}_1$, $\mathcal{S}_2$ and $\mathcal{S}_3$ be
these three sets. 

Each of the features can take values from the respective set of shapes. For a
particular experiment, let $v_1, v_2, v_3$ represent the values taken by Feature
1, Feature 2 and Feature 3 respectively. Then $v_1 \in \mathcal{S}_1$, $v_2 \in
\mathcal{S}_2$ and $v_2 \in \mathcal{S}_3$. Further, a feature label for an
instance of the time-series data can be represented as $\{v_1, v_2, v_3\}$.

Let the training data be represented by $\mathcal{D}$. For corruption, the
training dataset $\mathcal{D}$ is divided in a ratio of $\alpha:(1-\alpha)$. Let
the two divisions be represented as $\mathcal{D}_{\alpha}$ and
$\mathcal{D}_{1-\alpha}$. The feature label $\{v_{1\alpha}, v_{2\alpha},
v_{3\alpha}\}$ of every data point in $\mathcal{D}_\alpha$ is corrupted as:

\begin{center}
    $v_{1\alpha} \leftarrow v; v \sim \mathcal{S}_1 \slash \{v_{1\alpha}\}$  \\
    $v_{2\alpha} \leftarrow v; v \sim \mathcal{S}_2 \slash \{v_{2\alpha}\}$ \\
    $v_{3\alpha} \leftarrow v; v \sim \mathcal{S}_3 \slash \{v_{3\alpha}\}$ \\
\end{center}

$\mathcal{D}_\alpha$ with corrupted labels, denoted by $\hat{\mathcal{D}_\alpha}$ is then rejoined with
$\mathcal{D}_{1-\alpha}$ and shuffled to form the corrupted training dataset $\hat{\mathcal{D}}$:

\begin{center}
    $\hat{\mathcal{D}} = \{\hat{\mathcal{D}_\alpha}, \mathcal{D}_{1-\alpha}\}$
\end{center}

\section{Network Architectures}

\subsection{Chess Experiments}
The feature extractors use a 2D Convolutional Neural Network with ReLU
activations for embedding the images in case of both $NEUROLOG$ and
$NEUROLOG^{-}$.

$NEUROLOG$: We use the same model as used in the original implementation. The input to the network is a 28$\times$28 image and the output is a
vector of length 8 for classifying among the 8 classes (7 chess pieces + empty
class). The architecture is as given below:

\begin{itemize}
    \item Embedding Layers
    \begin{itemize}
        \item Conv2D (1, 6), 5$\times$5 filters, stride 1, padding 0 
        \item MaxPool2D, 2$\times$2 filters
        \item ReLU
        \item Conv2D (6,16), 5$\times$5 filters, stride 1, padding 0 
        \item MaxPool2D, 2$\times$2 filters
        \item ReLU
    \end{itemize}
    \item Classification MLP
    \begin{itemize}
        \item FC (256, 120)
        \item ReLU
        \item FC (120, 84)
        \item ReLU
        \item FC (84, 8)
    \end{itemize}
\end{itemize}

$NEUROLOG^{-}$: We use a network similar to the network used above with input and output layers slightly modified to support larger input and output sizes. The input to the network is an 84$\times$84 image, whereas the output is a vector of size 72 (9$\times$8). The first 8 outputs are used for classifying the first block of the chess board, next 8 outputs are used for classifying the second block and so on.

\begin{itemize}
    \item Embedding Layers
    \begin{itemize}
        \item Conv2D (1, 6), 5$\times$5 filters, stride 2, padding 0 
        \item MaxPool2D, 2$\times$2 filters
        \item ReLU
        \item Conv2D (6, 16), 5$\times$5 filters, stride 1, padding 0 
        \item MaxPool2D, 2$\times$2 filters
        \item ReLU
    \end{itemize}
    \item Classification MLP
    \begin{itemize}
        \item FC (1024, 512)
        \item ReLU
        \item FC (512, 256)
        \item ReLU 
        \item FC (256, 72)
    \end{itemize}
\end{itemize}

\subsection{Time Series Experiments}
The feature extractors use a 1D Convolutional Neural Network with ReLU
activations for embedding the images in case of both $NEUROLOG$ and
$NEUROLOG^{-}$.

$NEUROLOG$: The input to the network is a signal of length 50 which is obtained from separating the shapes. The output is a vector of size 9 for classifying from one of the 9 shapes. The architecture is as given below:

\vspace{0.25cm}
\begin{itemize}
    \item Embedding Layers
    \begin{itemize}
        \item Conv1D (1, 32), kernel size 8 filters, stride 1, padding 0 
        \item ReLU
        \item Conv1D (32, 32), kernel size 4, stride 1, padding 0 
        \item ReLU
        \item MaxPool1D, kernel size 2
    \end{itemize}
    \item Classification MLP
    \begin{itemize}
        \item FC (640, 256)
        \item ReLU
        \item Dropout 0.5 probability
        \item FC (256, 9)
    \end{itemize}
\end{itemize}

\vspace{0.25cm}

$NEUROLOG^{-}$: We use a network similar to the network used above with input and output layers slightly modified to support a larger input size. The input to the network is a 256 length signal. The output is a vector of size 9 (3$\times$3). The first 3 outputs are used for determining the shape taken by the first feature from $\mathcal{S}_1$ of the signal, the next 3 for the determining the shape taken by Feature 2 from $\mathcal{S}_2$ and last three for determining the shape taken by Feature 3 from $\mathcal{S}_3$

\vspace{0.25cm}
\begin{itemize}
    \item Embedding Layers
    \begin{itemize}
        \item Conv1D (1, 32), kernel size 8 filters, stride 1, padding 0 
        \item ReLU
        \item Conv1D (32, 32), kernel size 4, stride 1, padding 0 
        \item ReLU
        \item MaxPool1D, kernel size 2
    \end{itemize}
    \item Classification MLP
    \begin{itemize}
        \item FC (3936, 256)
        \item ReLU
        \item Dropout 0.5 probability
        \item FC (256, 9)
    \end{itemize}
\end{itemize}

\section{Deduction through Logical Theories}

In this section, we describe how the outputs of the neural feature extractor are used for predicting the features and the final class.

\subsection{Determining the Predicted Pieces in Chess Domain}
\begin{enumerate}
    \item From the outputs of the network, each of the 9 square are classified
    into belonging to one of the 8 states.
    \item The abductive feedback is collected for all the three classes
    \textit{safe}, \textit{mate}, \textit{draw}.
    \item The predicted set of features is compared to all the abductive proofs in the abductive feedbacks in an attempt to find an exact match.
    \item If a match is found the corresponding class is considered to be the
    predicted class. If a match is not found, the prediction is considered to be
    invalid. 
    
\end{enumerate}
 
\subsection{Determining the Predicted Conjunct in Time-Series Domain}

\begin{enumerate}
    \item Categorical probability distributions are calculated (using softmax)
    over the three sets $\mathcal{S}_1$, $\mathcal{S}_2$ and $\mathcal{S}_3$.
    Let's denote these distribution by $P_{\mathcal{S}_1}$, $P_{\mathcal{S}_2}$
    and $P_{\mathcal{S}_3}$.
    \item The probability of each of the conjuncts present in the theory is
    calculated by multiplying the probability of the component shapes. For eg.
    the probability of the conjunct $2 \land 0 \land 7$ in the theory given in
    Table \ref{table:eg_theory} is calculated as $P_{\mathcal{S}_1}(2) \times
    P_{\mathcal{S}_1}(0) \times P_{\mathcal{S}_1}(7)$ 
    \item The conjunct with the maximum probability is considered to be the
    predicted conjunct and the corresponding class is considered to be the
    predicted class.
\end{enumerate}

\section{Additional Training and Evaluation Details}

In this section we talk about some additional details about the experiments:

\subsection{Miscellaneous Details}
\begin{itemize}
    \item Cross Entropy Loss is used for pretraining the neural feature
    extractors in experiments on both chess and time-series domains.
    \item For the chess domain, we perform experiments on the 'BSV' scenario as described in NEUROLOG (Tsamoura, Hospedales, and Michael 2021) only. We DO NOT investigate the 'NGA' and 'ISK' scenarios.
    \item For chess experiments we use all 9000 data points provided in the train data for training and 2000 randomly sampled data points from the test data for evaluation.
\end{itemize}

\subsection{Hyperparameters}

Table \ref{table:details} lists further details and hyperparameters used in the experiments

\begin{table}[h]
\centering
\begin{tabular}{|c|c|c|c|}
\hline
                       & $NEUROLOG$ & \multicolumn{2}{|c|}{$NEUROLOG^{-}$} \\ 
\hline
                       &            &  Pretraining & Finetuning \\    
\hline
\multicolumn{4}{|c|}{Chess Experiments} \\
\hline
    Learning Rate      &    0.001   &   0.0001     &    0.00001   \\
\hline
    Optimizer          &    Adam    &   Adam       &    Adam      \\
\hline
    Train Epochs       &    15      &    40        &    15         \\
\hline
    Data Transforms      &   MinMax Scaling       &    MinMax Scaling      &  MinMax Scaling          \\
\hline 
    Batch Size         &     1       &    64       &     1         \\
\hline
\multicolumn{4}{|c|}{Time Series Experiments} \\
\hline
    Learning Rate      &    0.0001  &   0.0001     &    0.0001   \\
\hline
    Optimizer          &    Adam    &     Adam     &    Adam    \\
\hline
    Train Epochs       &    300     &      200     &    300        \\
\hline
    Data Transforms      &  Z Normalization &  Z Normalization &  Z Normalization \\
\hline 
    Batch Size         &      64      &     64         &    64          \\
\hline
    Upper Bound        &      8      &      8        &      8        \\
\hline
    Lower Bound        &        5    &      5        &      5         \\
\hline
\end{tabular}
\caption{Experimental Details for the Experiments. MinMax Scaling is done between [-0.5,0.5]}
\label{table:details}
\end{table}

\section{Additional Experiments and Results}
We further probe the effectiveness of $NEUROLOG^{-}$ on feature extractors pretrained with higher  levels of noise in the time-series domain. We increase the level of corruption from 0.1 to 0.6 in steps of 0.1. Table \ref{table:results} shows the results on the same.

\begin{table}[h]
\centering
\centering
\begin{tabular}{|cc|c|c|}
\hline
\textbf{Model} & & \textbf{Class acc.}& \textbf{Expl fid.} \\ 
\hline
\hline
$N$                                &                   &         100 (0.00)                &    100 (0.00)                          \\
\hline \hline
\multirow{6}{*}{$N_{1,\alpha}^{-}$}&   $\alpha = 0.1$  &         82.07 (6.59)          &        76.05 (10.23)                       \\

                                  &   $\alpha = 0.2$  &         81.48 (4.44)          &        76.08 (7.78)                       \\

                                  &   $\alpha = 0.3$  &         81.56 (4.61)              &        75.37 (10.28)                       \\
                                   
                                  &   $\alpha = 0.4$  &         76.80 (7.73)             &        67.70 (12.57)                       \\

                                  &   $\alpha = 0.5$  &         69.12 (7.24)             &        52.09 (11.44)                     \\

                                  &   $\alpha = 0.6$  &         62.71 (2.77)              &       29.88 (3.04)                      \\
\hline
$N_0^{-}$                         &                   &         75.79 (11.86)                &      36.81 (16.85)                   \\
\hline
\end{tabular}
\caption{Class accuracy and Explanatory Fidelity for decreasing quality of feature extractor (by increasing corruption $\alpha$).}
\label{table:results}
\end{table}

\begin{figure}[h]
     \centering
     \includegraphics[scale = 0.7]{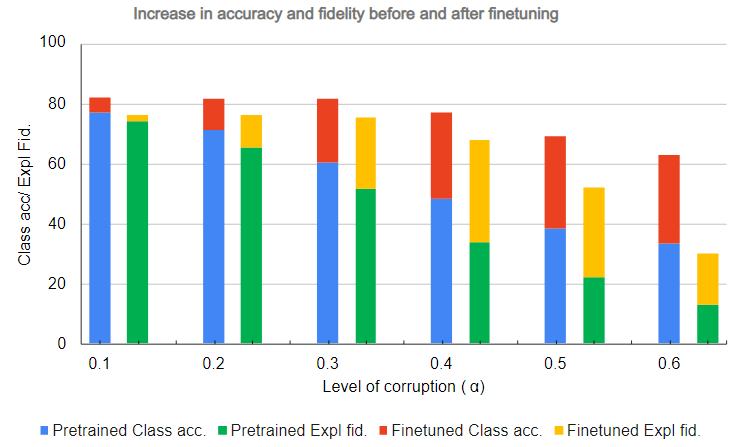}
     \caption{Pretrain and Finetune Class accuracy and Explanatory fidelity for different levels of corruption}
     \label{fig:bar_graph}
\end{figure}

Fig. \ref{fig:bar_graph} shows that the relative improvement using the semantic loss increases as the quality of feature extractors worsens, till a point. For the time series data, this maxima lies between $\alpha$ 0.4 and 0.5 with a gain of 28-33 \% in both class accuracy and exploratory fidelity.

\end{document}